%% file: iclr2021_workshop.tex
\documentclass{article}
\usepackage{iclr2021_workshop,times}

\iclrfinalcopy

\input{math_commands.tex}

\usepackage{hyperref}
\usepackage{url}

\usepackage{microtype}
\usepackage{graphicx}
\usepackage{tikz}
\usepackage{subfig}
\usepackage{algorithm}
\usepackage{algorithmic}

\usepackage{xcolor}
\usepackage{booktabs} 
\usepackage{amssymb} 
\usepackage{amsmath}
\usepackage{bm}

\usepackage{amsthm}
\usepackage{color}
\usepackage{amsfonts}

\newcommand{\norm}[1]{\left\lVert#1\right\rVert}

\newtheorem{definition}[]{Definition}[]

\usepackage{xspace}
\usepackage{multirow}
\newcommand{\Poincare}{Poincar\'e\xspace}

\title{A Geometry-Aware Algorithm to Learn Hierarchical Embeddings in Hyperbolic Space}

\author{Zhangyu Wang \thanks{The author is a PhD student at University of California, Santa Barbara. The corresponding email is \texttt{zhangyuwang@ucsb.edu}. The work is done during an internship at Alibaba Inc.} \\
Alibaba Inc. \\
Hangzhou, China 310000 \\
\texttt{zhangyu.wzy@alibaba-inc.com}\\
\AND
Lantian Xu \\
Carnegie Mellon University \\
Pittsburgh, PA, USA 15213 \\
\texttt{lxu2@andrew.cmu.edu}
\AND
Zhifeng Kong \\
University of California San Diego \\
La Jolla, CA, USA 92092 \\
\texttt{z4kong@eng.ucsd.edu}
\AND
Weilong Wang \\
Purdue University\\
West Lafayette, IN, USA, 47906\\
\texttt{wang4167@purdue.edu}
\AND
Xuyu Peng, Enyang Zheng \\
Alibaba Inc. \\
Hangzhou, China 310000 \\
\texttt{\{xijiu.pxy, enyang.zhengey\}@alibaba-inc.com}
}

\begin{document}

\maketitle

\begin{abstract}
Hyperbolic embeddings are a class of representation learning methods that offer competitive performances when data can be abstracted as a tree-like graph. However, in practice, learning hyperbolic embeddings of hierarchical data is difficult due to the different geometry between hyperbolic space and the Euclidean space. To address such difficulties, we first categorize three kinds of illness that harm the performance of the embeddings. Then, we develop a geometry-aware algorithm using a dilation operation and a transitive closure regularization to tackle these illnesses. We empirically validate these techniques and present a theoretical analysis of the mechanism behind the dilation operation. Experiments on synthetic and real-world datasets reveal superior performances of our algorithm.
\end{abstract}

\section{Introduction}\label{sec: introduction}
Learning data representation is important in machine learning as it provides a metric space that reveals or preserves inherent data structure \citep{mikolov2013distributed,pennington2014glove,bojanowski2017enriching,hoff2002latent,grover2016node2vec,perozzi2014deepwalk,nickel2011three,bordes2013translating,riedel2013relation}. Hyperbolic embeddings, a class of hierarchy representation methods, have shown competitive performances when data can be abstracted as a graph \citep{chamberlain2017neural, davidson2018hyperspherical,ganea2018ahyperbolic,gu2018learning,tifrea2018poincar}. 

In this work, we focus on the following embedding task. Let $\mathcal{D}$ be a dataset incorporated with a set of hierarchical relations represented as edges in a tree-like graph $\mathcal{G}$. The goal is to learn an embedding $\Theta$ in the \textit{hyperbolic} space by drawing positive and negative samples of edges from the graph such that $\Theta$ preserves the edge relationships, which are reflected by the order of similarity between data pairs. The formal problem statement is presented in Section \ref{sec: preliminaries}. 

Theoretically, hyperbolic space, such as the \Poincare Ball model, benefit from high representational power due to their negative curvatures \citep{Nickel2017,sala2018representation}. This observation has motivated research on solving real-world problems in hyperbolic space. For datasets with an \textit{observed} structure, hyperbolic space can embed the data and preserve the structure with arbitrarily low distortion \citep{Nickel2017,chamberlain2017neural,nickel2018learning,Ganea2018a,Chami2019a}. For datasets with a \textit{latent} structure, especially those obeying the power-law, hyperbolic space can provide a natural metric such that finer concepts are embedded into areas allowing more subtlety \citep{tifrea2018poincar,leimeister2018skip,le2019inferring}.

Despite the theoretical advantages of hyperbolic embeddings, learning such representation in practice is difficult. Specifically, the following fundamental difficulties have not been well-studied in the literature. (1) Many properties of the Euclidean space do not transfer to hyperbolic space. For example, the latter generally do not have the scale or shift-invariance in the sense of preserving similarity orders. (2) Many nice properties exclusive to hyperbolic space may improve learning. However, it is unclear how to design algorithms to effectively incorporate these properties. (3) Optimization in hyperbolic space is $(i)$ expensive due to a more sophisticated distance measure and $(ii)$ unstable because gradient descent is performed on hyperbolic manifolds. 

In this paper, we analyze these difficulties and provide a set of solutions to them.
First, we define bad cases as improper relationship between nodes and edges. We then categorize them into \textit{capacity illness}, \textit{inter-subtree illness}, and \textit{intra-subtree illness}. Formal definitions and intuitive visualizations are presented in Section \ref{sec: illness}. We present a theoretical analysis of local capacity, capacity illness, and their relationship in Section \ref{sec: local capacity}. 
We then develop an algorithm that reduces these illness in Section \ref{sec: algorithm}. The algorithm involves a \textit{dilation} operation during the learning process, adding transitive closure edges of data to positive samples, and a re-weighting strategy.
We conduct experiments on synthetic and real world datasets in Appendix \ref{sec: partial experiments}. The results show that our algorithm achieves superior performances under various evaluation metrics.

\section{Preliminaries}\label{sec: preliminaries}

A hyperbolic space $H^d$ is a 
$d$-dimensional Riemannian manifold with a constant negative sectional curvature. 
In this paper, we focus on the \Poincare ball model. Let $\mathcal{B} = \mathcal{B}^{d}$ denote the $d$-dimensional Poincare ball. The distance between any two points $B_1, B_2\in \mathcal{B}^{d}$ is defined as
\begin{equation}
    \label{Distance}
    d(B_1,B_2) = \mathrm{arcosh}\left(1+2\frac{\|u-v\|^2}{(1-\|u\|^2)(1-\|v\|^2)}\right),
\end{equation}
where $u$ and $v$ are the Euclidean vectors of $B_1$ and $B_2$. In the rest of the paper, we denote the \Poincare distance by $d(\cdot,\cdot)$ and the Euclidean distance by $\norm{\cdot}$.
Given a set of points $\mathcal{V}=\{x_i\}_{i=1}^n$ and the relation set $\mathcal{E}\subset [n]^2$, the goal is to learn an embedding $f: \mathcal{V}\to \mathcal{B}$ that preserves the inherent structure. To achieve this goal, we define and minimize the following loss function $\mathcal{L}$. For $(i,j)\in\mathcal{E}$, define $\mathcal{N}(x_i,x_j)$ as the set of negative samples of $(i,j)$. Let $\Theta=\{\theta_i\}_{i=1}^n$, where each $\theta_i\in\mathcal{B}^d$ is the embedding of $x_i$. Define $d(x_i,x_j)=d(\theta_i,\theta_j)$. Then, the loss function is defined as
\begin{equation}\label{eq: loss}
    \mathcal{L}(\Theta) = -\sum_{(i,j)\in \mathcal{E}}
    \log\frac{e^{-d(x_i,x_j)}}{\sum_{x'\in \mathcal{N}(x_i,x_j)\cup \{x_j\}}e^{-d(x_i,x')}} = -\sum_{(i,j)\in \mathcal{E}}\mathcal{L}_{i,j}(\Theta).
\end{equation}
This objective can be optimized via Riemannian gradient descent \citep{Nickel2017}.

\section{Illness}\label{sec: illness}
In this section, we formally define illness that harms the performance of hyperbolic embeddings and is hard to optimize. Let $\overrightarrow{AB}$ be a ground-truth edge in $G$, and $\overrightarrow{AB'}$ be the inferred edge from the hyperbolic embeddings, where $B'\neq B$. We call this situation \textit{illness with respect to $A$}. 
Let $C$ be the nearest common ancestor of $B$ and $B'$. We categorize three kinds of illness according to the pairwise relationships among $B$, $B'$, and $C$. Formally, we define capacity illness, intra-subtree illness, and inter-subtree illness in \textbf{Definition} \ref{def: illness}.

\begin{definition}[Categories of Illness] \label{def: illness}
    We define the illness to be capacity illness if $B$ is the parent of $B'$. We define the illness to be an intra-subtree illness if $B$ is the ancestor but not the parent of $B'$. We define the illness to be an inter-subtree illness if $C\neq B$.
\end{definition}

It is straightforward to see that the union of capacity illness and intra-subtree illness are exactly situations where $C=B$. Therefore, the above three kinds of illness are a partition of all illness. We visualize these three kinds of illness in Figure \ref{fig: illness} in the appendix.

\section{Local Capacity}\label{sec: local capacity}

We define local capacity below and theoretically relate it to capacity illness.

\begin{definition}[Local Capacity] \label{def: local capacity}
    Given a geodesic space $(\mathcal{X},d)$ and a geodesic ball $\mathcal{S}_r$ centered at $A\in \mathcal{X}$ with radius $r$. The local capacity of $(A,r)$ is defined as 
    \begin{equation}
    \label{lcapacity}
        \max\left\{|\mathcal{C}|:\mathcal{C}\in\mathcal{S}_r; \forall C_1,C_2\in\mathcal{C}, C_1\neq C_2, d(C_1,C_2) > d(C_1, A)\vee d(C_2, A)\right\}.
    \end{equation}
\end{definition}

Given a geodesic space $(\mathcal{X},d)$ and a geodesic ball $\mathcal{S}_r$ centered at $A\in \mathcal{X}$ with radius $r$. The local capacity of $(A,r)$ is defined as 
    \begin{equation}
    \label{lcapacity}
        \max\left\{|\mathcal{C}|:\mathcal{C}\in\mathcal{S}_r; \forall C_1,C_2\in\mathcal{C}, C_1\neq C_2, d(C_1,C_2) > d(C_1, A)\vee d(C_2, A)\right\}.
    \end{equation}
    
For not very small $r$ and large $d$ we have the following bounds:
\begin{equation}\label{eq: local capacity}
    2^d e^{\frac{dr}{2}} \gtrapprox \mathcal{A}(d, \theta_r) \gtrapprox \sqrt{2\pi}\log\frac{2}{\sqrt{3}}\cdot d^{\frac32}\cdot 2^{1-d} e^{\frac{d-1}{2}r},
\end{equation}
where $\theta_r=\arcsin\left(1/(2\cosh(r/2))\right)$. Full derivations are in Appendix \ref{appendix: illness}.

\section{The Algorithm}\label{sec: algorithm}
In this section, we build a geometry-aware algorithm (Algorithm \ref{alg:G-Aalgorithm}) targeting the three categories of illness 
by proposing the dilation operation and the transitive closure regularization.

\begin{algorithm}[tb]
\caption{Geometry-Aware Algorithm}
\label{alg:G-Aalgorithm}
\begin{algorithmic}
   \FOR{$i=1$ {\bfseries to} $N_{\mathrm{epoch}}$}
   \STATE Compute local capacity according to \eqref{eq: local capacity}
   \STATE \textbf{if} local capacity is not sufficient \textbf{then} perform the dilation operation in \eqref{eq: dilation}
   \STATE \textbf{end if}
   \STATE \textbf{if} $i\leq N_{\mathrm{tc}}$ \textbf{then} loss $\leftarrow \mathcal{L}_{\mathrm{tc}}(\Theta)$ according to \eqref{eq: loss + transitive closure}
    \STATE \textbf{else} loss $\leftarrow \mathcal{L}(\Theta)$ according to \eqref{eq: loss}
    \STATE \textbf{end if}
   \STATE Apply Riemannian gradient descent over loss
   \ENDFOR
\end{algorithmic}
\end{algorithm}

\paragraph{Dilation.} We define a mapping $g: \mathcal{B}\to \mathcal{B}$ as a $k$-dilation if for any $A\in \mathcal{B}$:
\begin{equation}\label{eq: dilation}
    d(O,g(A)) = k\cdot d(O,A).
\end{equation}
Notably, $g$ can be computed explicitly. For instance, a $2$-dilation can be formulated as $g(A) = \frac{2}{1+\norm{A}^2}A$. 
The dilation operation rescales the embedded structure so that each point is pushed to a location with sufficient local capacity. Given $A\in \mathcal{B}$ with degree $k$, 
this operation helps increase the distance between $A$ and its $k$-nearest neighbor($r_A$), thus increase the local capacity of $(A, r_A)$.

\paragraph{Transitive closure regularization.} It contains the following two operations.

\textit{Adding transitive closure edges}. The transitive closure edges $\mathcal{E}_{\mathrm{tc}}$ are edges between nodes and their non-parent ancestors. These edges are also considered as positive samples in addition to $\mathcal{E}$ in the objective in \eqref{eq: loss}. The purpose of adding these auxiliary edges is to push the subtrees apart so they are less likely to overlap in the \Poincare ball.

\textit{Re-weighting.} We modify the weights of transitive closure edges to prevent overfitting in early (the first $N_{\mathrm{tc}}$) epochs. Let $\eta_{\mathrm{tc}}$ be a real number between 0 and 1. Then, the objective becomes
\begin{equation}\label{eq: loss + transitive closure}
\mathcal{L}_{\mathrm{tc}}(\Theta) = \mathcal{L}(\Theta) + \eta_{\mathrm{tc}}\sum_{(i,j)\in \mathcal{E}_{\mathrm{tc}} }
    \mathcal{L}_{i,j}(\Theta).
\end{equation}

It is noteworthy that these operations are not admissible in the Euclidean space, where the local capacity of any $(A,r)\in\mathbb{R}^d\times \mathbb{R}$ is a constant with respect to $d$.

\section{Experiments}\label{sec: partial experiments}

\begin{figure*}[!t]
    \centering
    \subfloat[The baseline \Poincare embedding algorithm]{
        \label{fig:figure2a}
        \begin{minipage}[t]{0.49\linewidth}
            \centering
            \begin{tikzpicture}
                \node (image) at (4,0) {
                \includegraphics[width=0.5\linewidth]{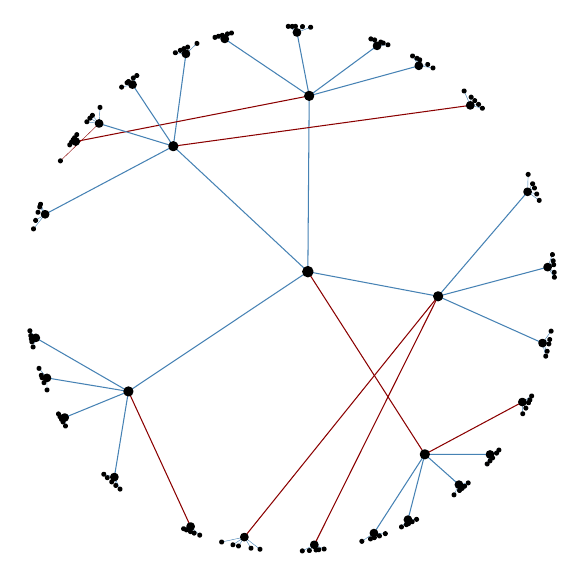}};
                \node (image) at (0,1) {
                \includegraphics[trim=20 190 180 10,clip,width=0.3\linewidth]{figures/figure-2a.pdf}
                };
                \draw[gray, thick] (2.4,0.5) rectangle (3.6,1.7);
                \draw[gray, thick] (-1.1,-0.4) rectangle (1.4,2.1);
                \draw[gray, thick, dashed] (2.4,0.5) -- (1.4,-0.4);
                \draw[gray, thick, dashed] (2.4,1.7) -- (1.4,2.1);
                
            \end{tikzpicture}
        \end{minipage}}
    \subfloat[Our algorithm]{
        \label{fig:figure2b}
        \begin{minipage}[t]{0.49\linewidth}
            \centering
            \begin{tikzpicture}
                \node (image) at (4,0) {
                \includegraphics[width=0.5\linewidth]{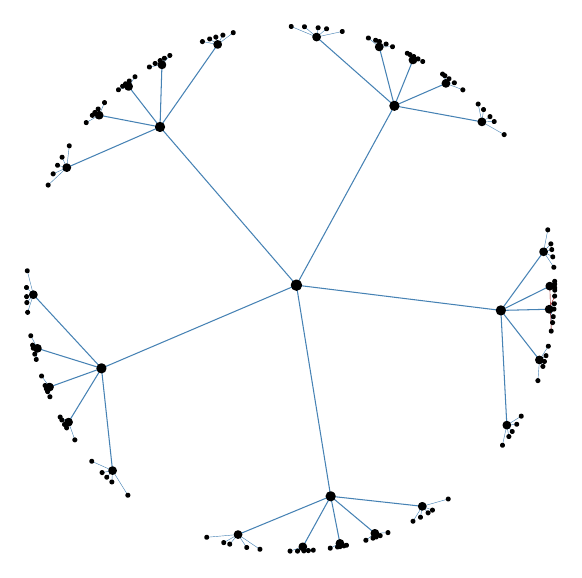}};
                \node (image) at (0,1) {
                \includegraphics[trim=10 185 190 15,clip,width=0.3\linewidth]{figures/figure-2b.pdf}};
                \draw[gray, thick] (2.35,0.45) rectangle (3.55,1.65);
                \draw[gray, thick] (-1.1,-0.4) rectangle (1.4,2.1);
                \draw[gray, thick, dashed] (2.35,0.45) -- (1.4,-0.4);
                \draw[gray, thick, dashed] (2.35,1.65) -- (1.4,2.1);
            \end{tikzpicture}
        \end{minipage}}
\caption{Visualizations of two-dimensional embeddings of a synthetic balance tree (156 nodes, 155 edges) learned by the baseline \Poincare embedding algorithm in \citet{Nickel2017} and our geometry-aware algorithm, respectively. Both algorithms are trained for 3000 epochs. The lines refer to ground-truth edges and the points refer to the learned hyperbolic embeddings. The red lines indicate bad cases where the embeddings fail to reconstruct these ground-truth edges.}
\label{fig:figure2}
\end{figure*}

\begin{figure*}[t]
\centering
\subfloat{\includegraphics[width=0.1\columnwidth]{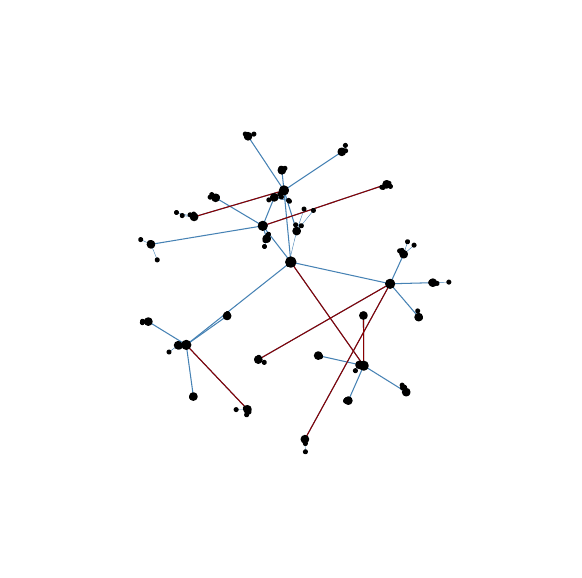}}\quad
\subfloat{\includegraphics[width=0.1\columnwidth]{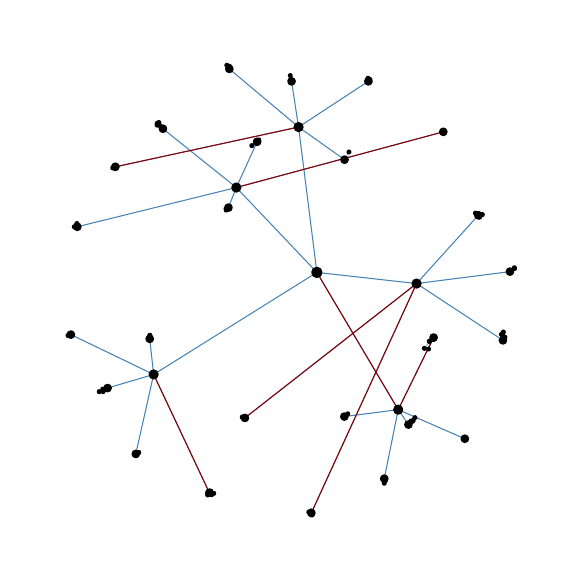}
}\quad
\subfloat{\includegraphics[width=0.1\columnwidth]{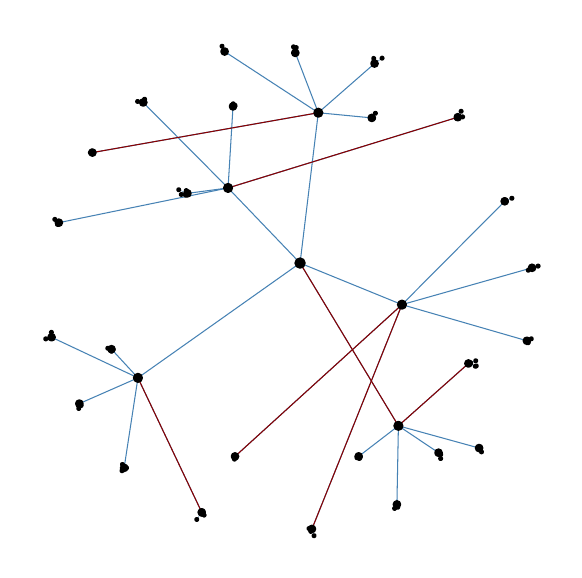}}\quad
\subfloat{\includegraphics[width=0.1\columnwidth]{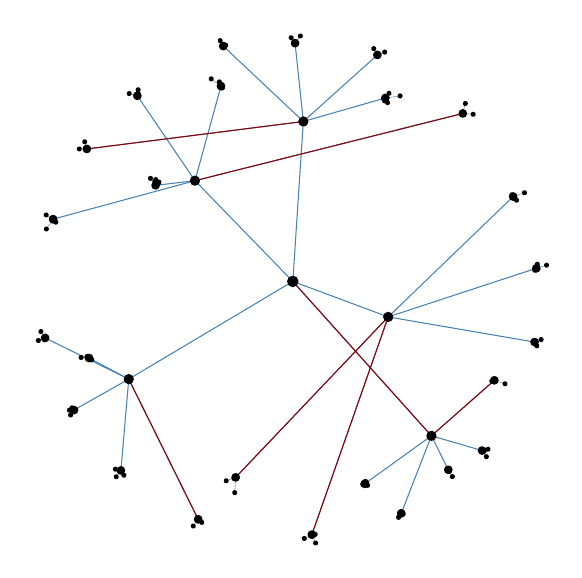}}\quad
\subfloat{\includegraphics[width=0.1\columnwidth]{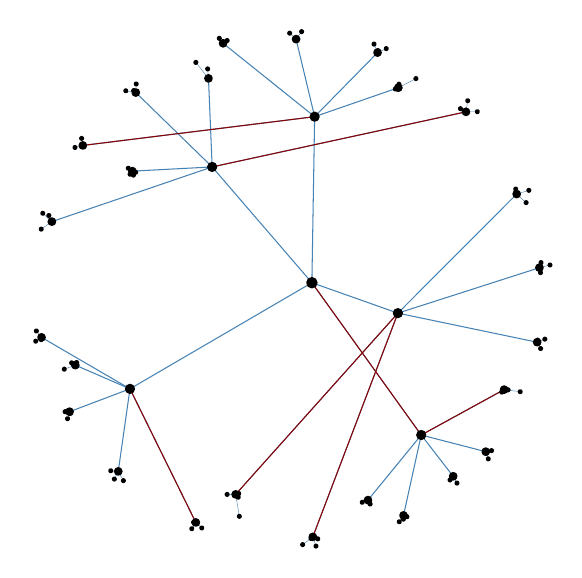}}\quad
\subfloat{\includegraphics[width=0.1\columnwidth]{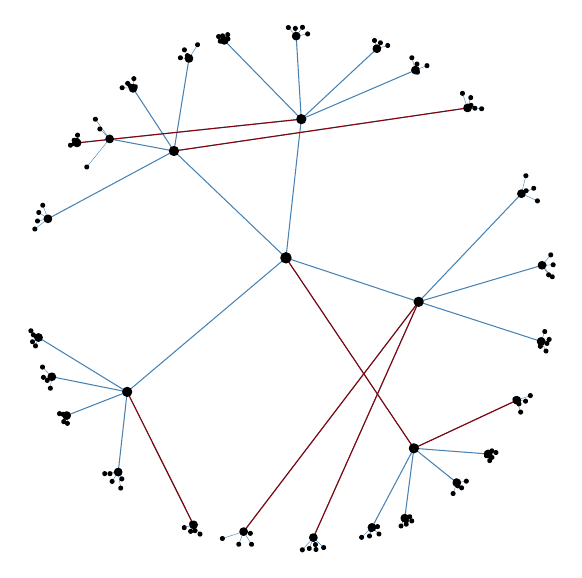}}\quad
\subfloat{\includegraphics[width=0.1\columnwidth]{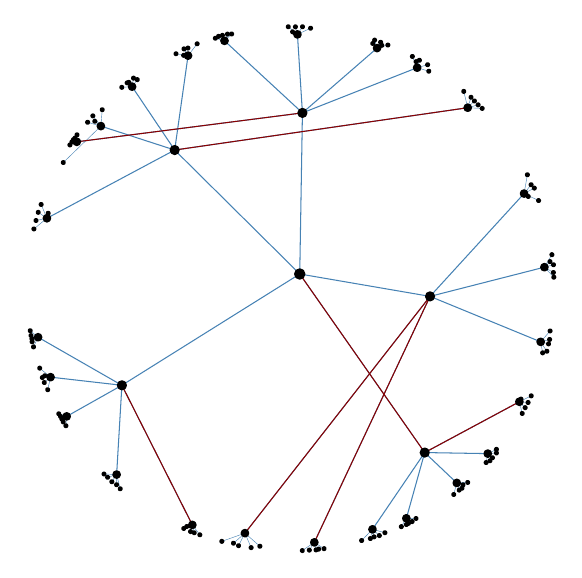}}\quad
\subfloat{\includegraphics[width=0.1\columnwidth]{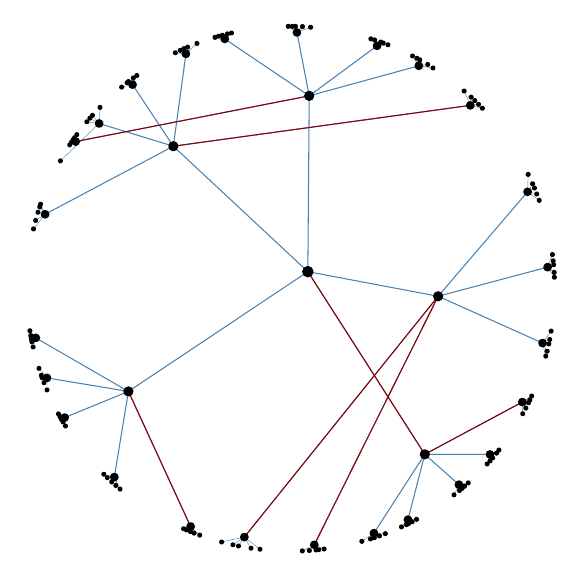}}\quad
\\
\subfloat{\includegraphics[width=0.1\columnwidth]{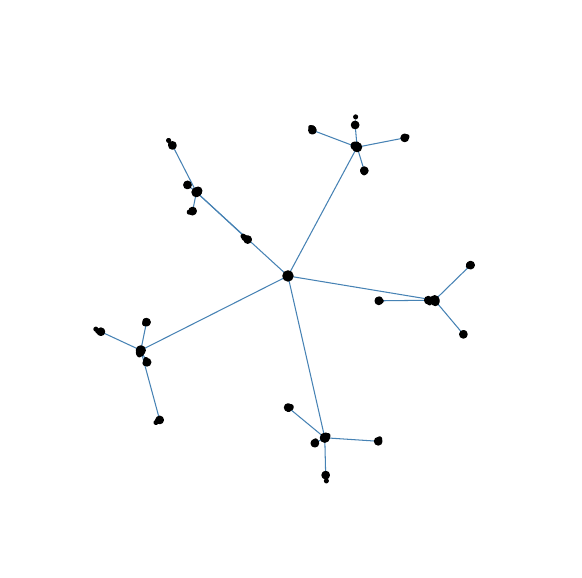}}\quad
\subfloat{\includegraphics[width=0.1\columnwidth]{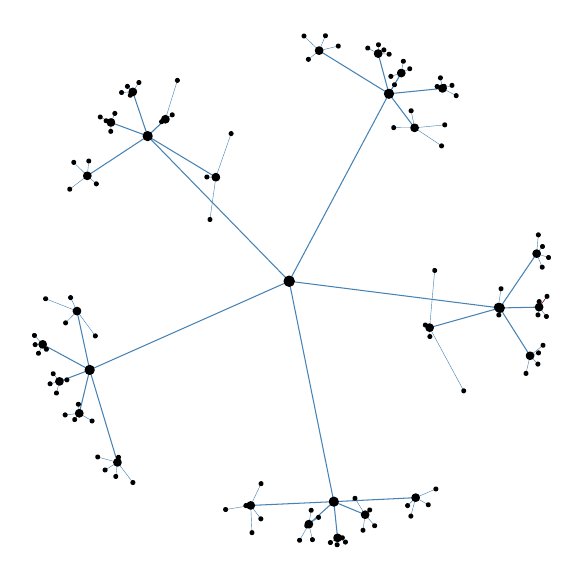}
}\quad
\subfloat{\includegraphics[width=0.1\columnwidth]{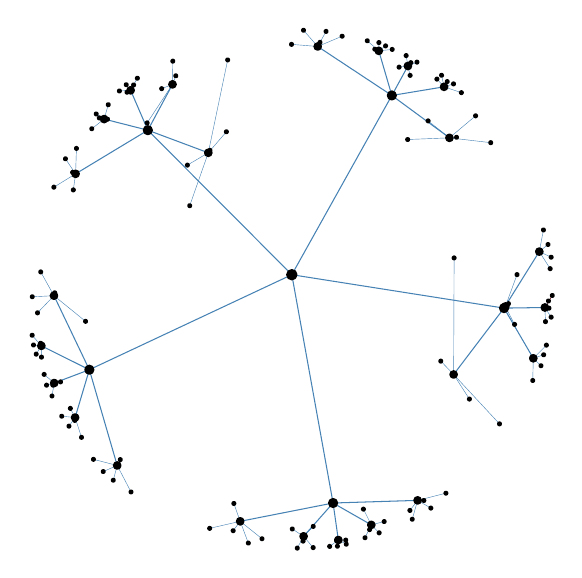}}\quad
\subfloat{\includegraphics[width=0.1\columnwidth]{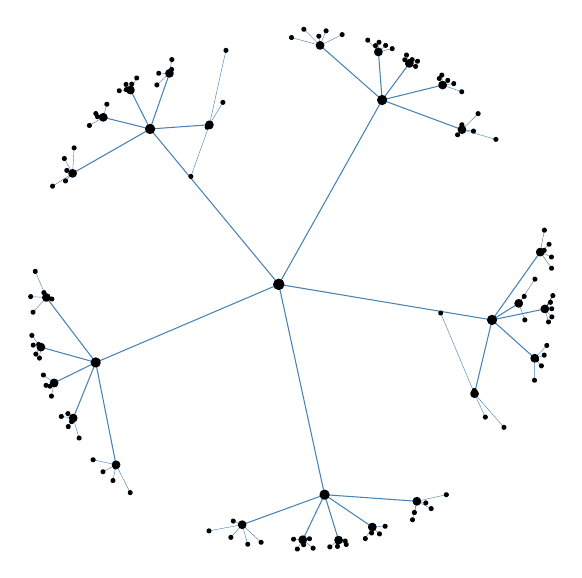}}\quad
\subfloat{\includegraphics[width=0.1\columnwidth]{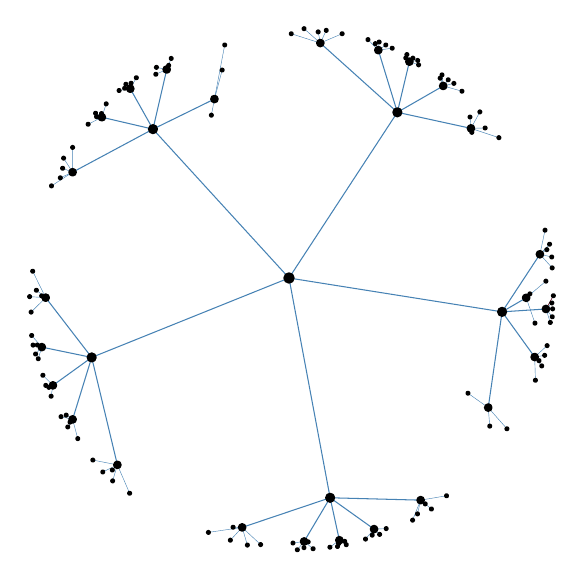}}\quad
\subfloat{\includegraphics[width=0.1\columnwidth]{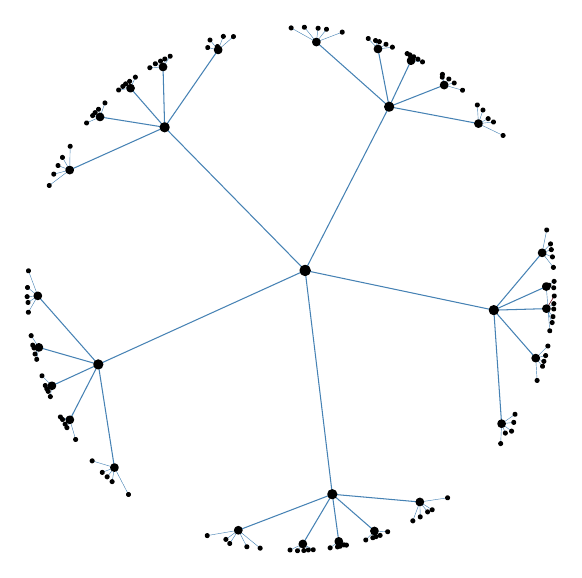}}\quad
\subfloat{\includegraphics[width=0.1\columnwidth]{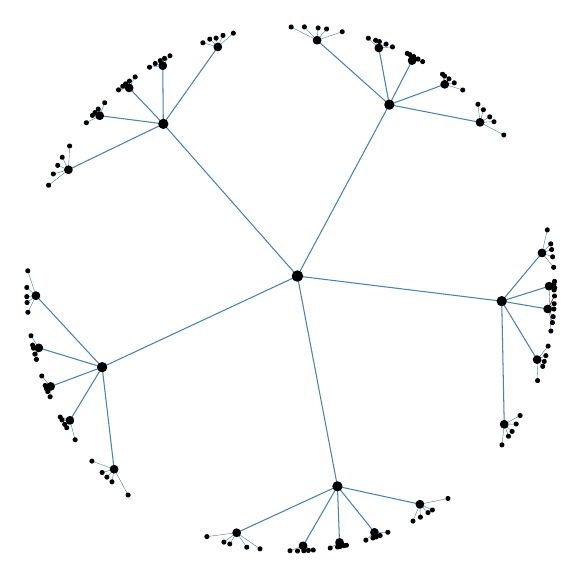}}\quad
\subfloat{\includegraphics[width=0.1\columnwidth]{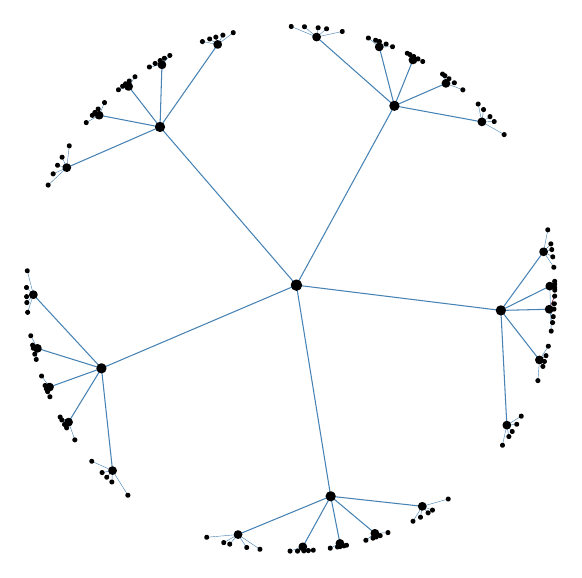}}
\caption{Visualization of the learning process of two-dimensional \Poincare embeddings with (1) the baseline algorithm in the upper row, and (2) the geometry-aware algorithm in the lower row. The dataset, baseline model, and plotting settings are identical as in Figure \ref{fig:figure2}. We plot intra-subtree and inter-subtree illness consistently existing throughout the entire 3000 epochs.}
\label{fig: syn process}
\end{figure*}

We compare our algorithm to the baseline model \citep{Nickel2017} on the synthetic dataset in Figure \ref{fig:figure2} and Figure \ref{fig: syn process}. Our method not only achieves perfect MAP (\textbf{0.998}) but also yields better reconstructed geometry. We do extensive experiments on multiple real-world datasets of various scales and characteristics in Appendix \ref{sec: experiments}. Results show our algorithm consistently outperform the baseline algorithms \citep{Nickel2017, nickel2018learning}, especially on extremely bushy datasets.

\section{Conclusion}\label{sec: conclusion}
In this paper, we analyze three categories of illness and develop a geometry-aware algorithm that targets at reducing these illnesses and improving performance. Our algorithm shows superior performance over baseline models on both synthetic and real world datasets.

\bibliography{iclr2021_workshop}
\bibliographystyle{iclr2021_workshop}

\newpage
\appendix
\input{appendix}

\end{document}

%% file: math_commands.tex
\usepackage{amsmath,amsfonts,bm}

\def\eqref#1{equation~\ref{#1}}

\def\1{\bm{1}}

\DeclareMathAlphabet{\mathsfit}{\encodingdefault}{\sfdefault}{m}{sl}
\SetMathAlphabet{\mathsfit}{bold}{\encodingdefault}{\sfdefault}{bx}{n}



%% file: appendix.tex

\section{Related Work}\label{sec: related work}
Learning in hyperbolic space is initially proposed by \citet{Nickel2017}. It is the most related work to our paper. Their method outperforms the Euclidean counterpart in low dimensions as to the task of learning embeddings for edge reconstruction. However, since their algorithm is directly adapted from the Euclidean space, it does not naturally leverage potentially useful geometrical properties of hyperbolic space (see Section \ref{sec: preliminaries}). As a consequence, there remain many bad cases even after convergence (see Figure \ref{fig:figure2a}).

A series of work directly incorporate properties of hyperbolic space via optimization \citep{wilson2018gradient,bonnabel2013stochastic,absil2009optimization,afsari2013convergence}. Specifically, \citet{nickel2018learning} conduct training in the Lorentz space with a closed-form expression of the geodesics on the hyperbolic manifold. However, since the learning objective \eqref{eq: loss} is highly non-convex, obtaining more accurate gradients does not completely solve the problem. 

Another group of work either implement hyperbolic versions of commonly used neural network modules \citep{ganea2018bhyperbolic,gulcehre2018hyperbolic,chami2019hyperbolic}, or design models specifically tailored for hyperbolic space \citep{vulic2017specialising,le2019inferring,cho2019large,leimeister2018skip,weber2020robust,chami2019low}. These methods are task-specific and thus expensive to deploy in downstream applications. 

Apart from the above learning approaches, \citet{sala2018representation} presents a combinatorial algorithm that achieves better performance than \citet{Nickel2017, nickel2018learning} with even lower dimensions. The core idea is to extend the 2-dimensional results of \citet{sarkar2011low} to arbitrary dimensions. However, this algorithm suffers from three vital weaknesses: (1) it requires complete information of the graph; (2) it is sensitive to addition/removal of data; and (3) most critically, it involves discrete operations and thus does not have gradients. Therefore, in scenarios where complete information is unavailable, the graph dynamically changes, or joint learning is needed, this approach does not suffice.


In this paper, we endorse the importance of leveraging geometrical properties in learning unsupervised hyperbolic embeddings. Based on this intention, we develop a geometry-aware algorithm that improves embedding performances, which, to the best of our knowledge, is original.

\section{Illness}\label{appendix: illness}

\begin{figure}[!h]
\centering
\includegraphics[width=0.8\columnwidth]{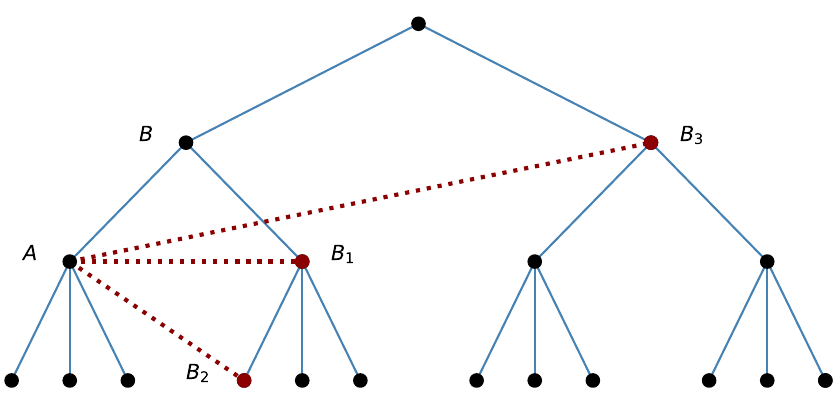}
\caption{Illustration of the three categories of illness. $A$ is the source node and $B$ is the ground-truth target node. It is called (1) \textit{capacity illness} if $A$ connects to $B_1$, (2) \textit{intra-subtree illness} if $A$ connects to $B_2$, and (3) \textit{inter-subtree illness} if $A$ connects to $B_3$.}
\label{fig: illness}
\end{figure}

\section{Local Capacity}\label{appendix: local capacity}

According to \textbf{Definition} \ref{def: local capacity}, for $A\in\mathcal{V}$ and a radius $r$, if $|\{C: C\text{ is a child of }A, d(A,C)\leq r\}|$ exceeds the local capacity of $(A,r)$, then capacity illness must exist. To obtain bounds on local capacity, we first bound the $r$-packing number of a geodesic sphere centered at point $A\in\mathcal{B}$ with radius $r$. 

First, the packing number of the geodesic sphere centered at the origin $O$ is the same as that of $A$. To see why we use a result from \citet{IntroHG}:
\begin{equation}
    \textbf{Isom}(\mathcal{B}) = \textbf{M\"ob}(\mathcal{B}).
\end{equation}
That is, the group of isometries from $\mathcal{B}$ to itself coincide with the group of all M\"obius transformations preserving $\mathcal{B}$. In the \Poincare ball model, \textbf{M\"ob}($\mathcal{B}$) is generated by inversions in generalized spheres $\mathcal{S}'$ such that $\mathcal{S}'\bot \partial \mathcal{B}$, therefore once we extend $\overrightarrow{O A}$ to $C$ with $\norm{OC}^2-1 = \norm{OC}\cdot\norm{CA}$, then by taking the restriction in $\mathcal{B}$ of the inversion in generalized sphere $\mathcal{S}'$ centered at $C$ we get an isometry from $\mathcal{B}$ to itself which maps $A$ to $O$. Specifically, it is an isometry between any geodesic sphere centered at $A$ and the geodesic sphere centered at the origin $O$ (with the same radius $r$).

Then, we compute the $r$-packing number of the geodesic sphere $S_r$ centered at $O$ with \Poincare radius $r$. For $B_1,B_2\in S_r$, let $u=\overrightarrow{OB_1},v=\overrightarrow{OB_2}$ and $r$ be the \Poincare norm of $u$. Then, as long as the angle $\theta$ between $u$ and $v$ satisfies
\begin{equation}
    \theta \geq \theta_r = 2 \arcsin\left(\frac{1}{2\cosh(r/2)}\right),
\end{equation}
we have $d(B_1,B_2) \geq d(B_i,O)$, $i=1,2$. For not very small $r$, $\theta_r \approx 2e^{-r/2}$. Then, the $r$-packing problem is equivalent to evaluating the size of the largest spherical code of angle $\theta_r$ in dimension $d$, defined as $\mathcal{A}(d,\theta_r)$. According to \citet{KissingN}, we have
\begin{equation}
    \mathcal{A}(d,\theta)\geq(1+o(1))\frac{c_\theta\cdot d}{s_d(\theta)},
\end{equation}
where $c(\theta) = \log\frac{\sin^2(\theta)}{\sqrt{(1-\cos\theta)^2(1+2\cos\theta)}}\approx \log\frac{2}{\sqrt{3}}$ for small $\theta$, $s_d(\theta) = (1+o(1))\frac{\sin^{d-1}\theta}{\sqrt{2\pi d}\cdot \cos\theta}$. Specifically, when $d\leq 16$ with small $\theta$, \citet{OldSC} provides a better bound:
\begin{equation}
    \mathcal{A}(d,\theta) \geq \frac{1}{s_d(\theta)} = (1+o(1))\frac{\sqrt{2\pi d}\cdot \cos\theta}{\sin^{d-1}\theta}.
\end{equation}
To sum up, for not very small $r$ we have the following lower bounds under different dimensions:
\begin{equation}\label{eq: local capacity lower bound}
    \mathcal{A}(d, \theta_r) \gtrapprox \left\{
    \begin{array}{cl}
        \pi e^{\frac{r}{2}} &  d = 2 \\
        \sqrt{2\pi d}\cdot 2^{1-d} e^{\frac{d-1}{2}r} &   3\leq d\leq 16 \\
        \sqrt{2\pi}\log\frac{2}{\sqrt{3}}\cdot d^{\frac32}\cdot 2^{1-d} e^{\frac{d-1}{2}r} &   d\geq 17
    \end{array}
    \right..
\end{equation}
As for the upper bound, according to \citet{UpperKN},
\begin{equation}
    \mathcal{A}(d,\theta)\leq e^{\phi(\theta) d(1+o(1))},
\end{equation}
where $\phi(\theta)>-\log \sin\theta$ is a certain function. Therefore, 
\begin{equation}\label{eq: local capacity upper bound}
    \mathcal{A}(d, \theta_r) \lessapprox \left\{
    \begin{array}{cl}
        \pi e^{\frac{r}{2}} &  d = 2 \\
        2^d e^{\frac{dr}{2}} &   d \geq 3
    \end{array}
    \right..
\end{equation}
Note that by considering the extension of radius, one can define local capacity on geodesic balls instead of spheres(\ref{lcapacity}), which leads to the same conclusion.

\newpage

\section{Experimental Settings and Overview}\label{sec: experiments}

We apply our algorithm to both a synthetic dataset and real-world datasets on the graph reconstruction task. We evaluate the performance by mean average precision (MAP) and Mean Rank (MR) defined below. For $A\in \mathcal{V}$ with degree $\mathrm{deg}(A)$ and neighborhood $\mathcal{N}_A = \left\{B_1,...B_{\mathrm{deg}(A)}\right\}$, let $R_{A,B_i}$ be the smallest subset of $\mathcal{V}$ containing $B_i$ and all points closer to $A$ than $B_i$. Then, the MAP is defined as
\begin{equation}
    \text{MAP}(f) = \frac{1}{|\mathcal{V}|}\sum_{A\in \mathcal{V}}\frac{1}{\mathrm{deg}(A)}\sum_{i=1}^{|\mathcal{N}_A|}\frac{|\mathcal{N}_A\cap R_{A,B_i}|}{|R_{A,B_i}|}.
\end{equation}
and the MR is defined as
\begin{equation}
    \text{MR}(f) =  \frac{1}{|\mathcal{V}|}\sum_{A\in \mathcal{V}}\sum_{i=1}^{|\mathcal{N}_A|}\left(| R_{A,B_i}|-i\right).
\end{equation}

In addition, we report the number of three kinds of illness defined in \textbf{Definition} \ref{def: illness} after the algorithm converges.

We run baseline algorithms \citep{Nickel2017, nickel2018learning} and our algorithm on a synthetic dataset, Yelp Challenge (Tree) (see Table \ref{table: yelp}), WordNet Verbs (Tree) (see Table \ref{table:wordnet verbs}), WordNet Nouns (Tree) (see Table \ref{table:wordnet nouns}), Commodity Catalog (Tree) (see Table \ref{table:commodity catalog}) and WordNet Nouns (Closure) (see Table \ref{table:wordnet nouns closure}). We report the reconstruction Mean Rank, MAP, number of capacity errors, number of intra-subtree errors and number of inter-subtree errors respectively.

A key point to notice is that we focus on tree datasets instead of general DAG or transitive closures of trees. In terms of the objective (reducing MAP and MR), learning a tree structure is much harder because the size of the neighborhood set is as few as one. Experiments validate this statement: we apply the baseline algorithm \citep{Nickel2017} to the same WordNet Noun Hierarchy dataset \citep{Nickel2017} where the transitive closure edges are removed. The performances in terms of MAP and MR significantly drop compared to the numbers reported in \citep{Nickel2017} (See Table \ref{table:wordnet nouns}).

\newpage

\section{Synthetic Dataset Experiments}\label{sec: synthetic}

\begin{figure*}[h!]
\centering
\subfloat[\label{fig:figure5a}Capacity illness]{\includegraphics[width=0.3\linewidth]{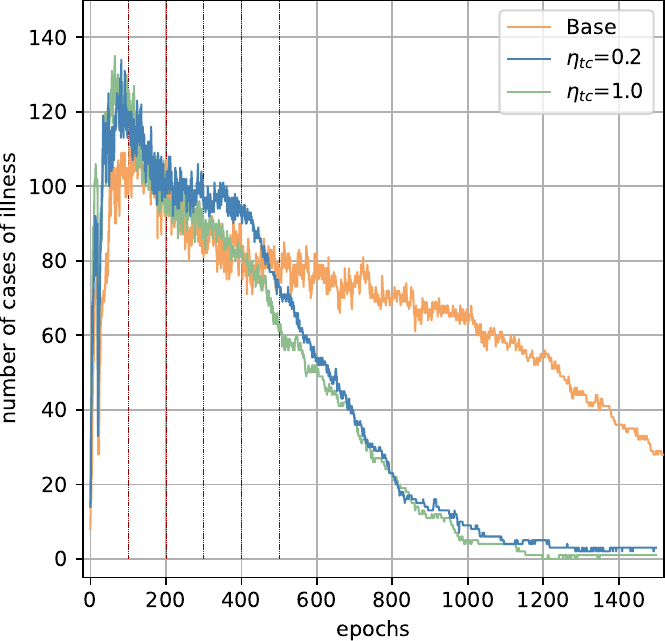}}\quad
\subfloat[\label{fig:figure5b}Intra-subtree illness]{\includegraphics[width=0.3\linewidth]{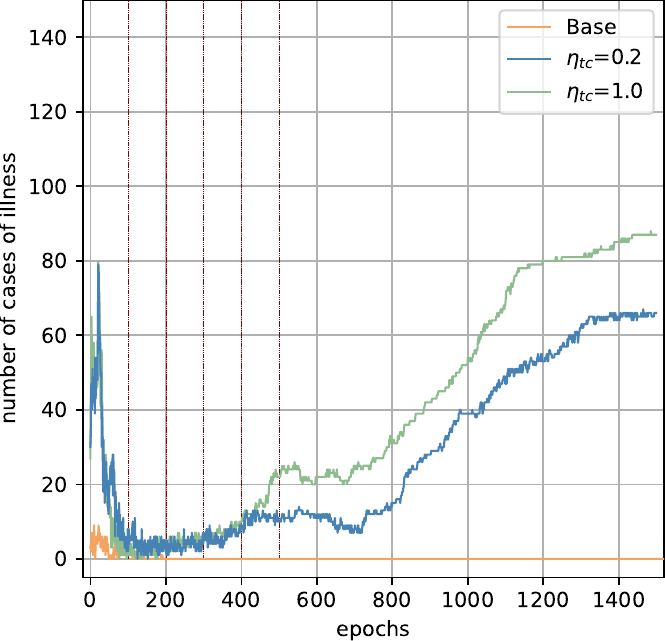}}\quad
\subfloat[\label{fig:figure5c}Inter-subtree illness]{\includegraphics[width=0.3\linewidth]{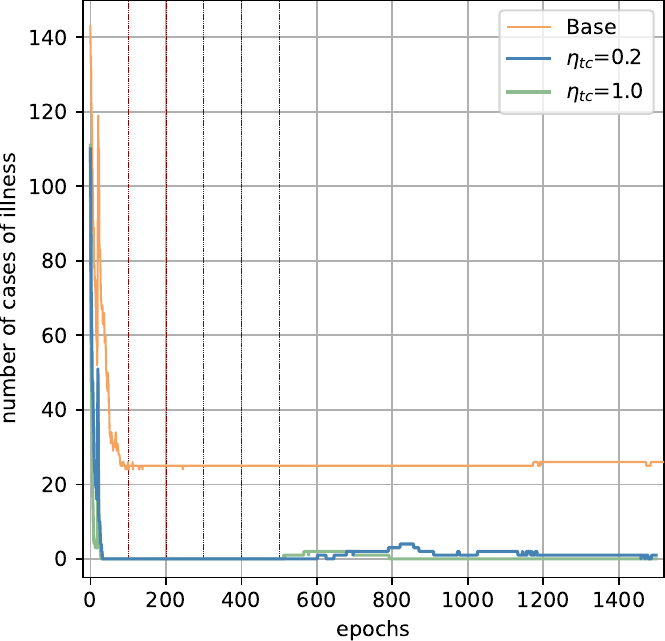}}
\caption{Number of different kinds of illness under different $\eta_{tc}$. Base denotes the baseline algorithm.}
\label{fig: syn tc weight}
\end{figure*}

\begin{figure*}[h!]
\centering
\subfloat[\label{fig:figure6a}Capacity illness]{\includegraphics[width=0.3\linewidth]{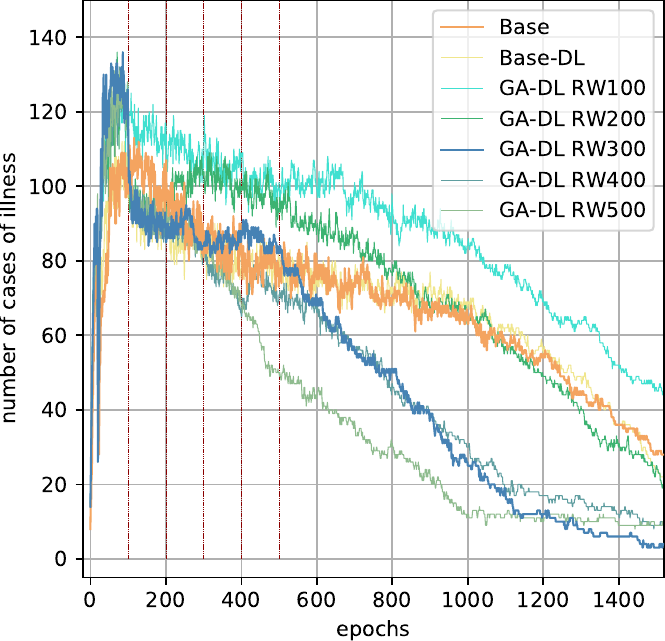}}\quad
\subfloat[\label{fig:figure6b}Intra-subtree illness]{\includegraphics[width=0.3\linewidth]{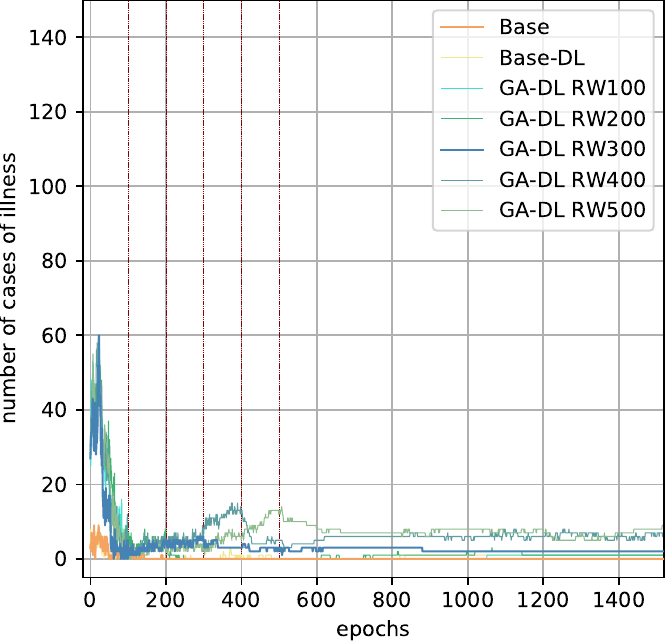}}\quad
\subfloat[\label{fig:figure6c}Inter-subtree illness]{\includegraphics[width=0.3\linewidth]{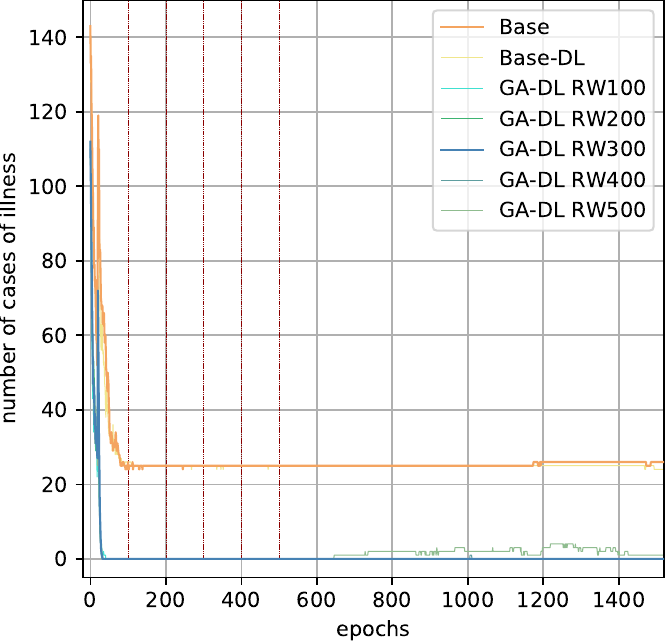}}\quad
\subfloat[\label{fig:figure6d}log MR]{\includegraphics[width=0.3\linewidth]{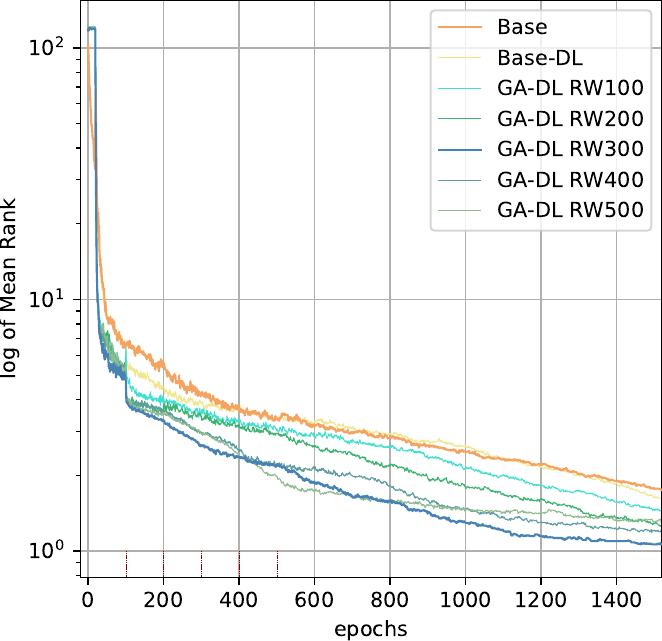}}\quad
\subfloat[\label{fig:figure6e}MAP]{\includegraphics[width=0.3\linewidth]{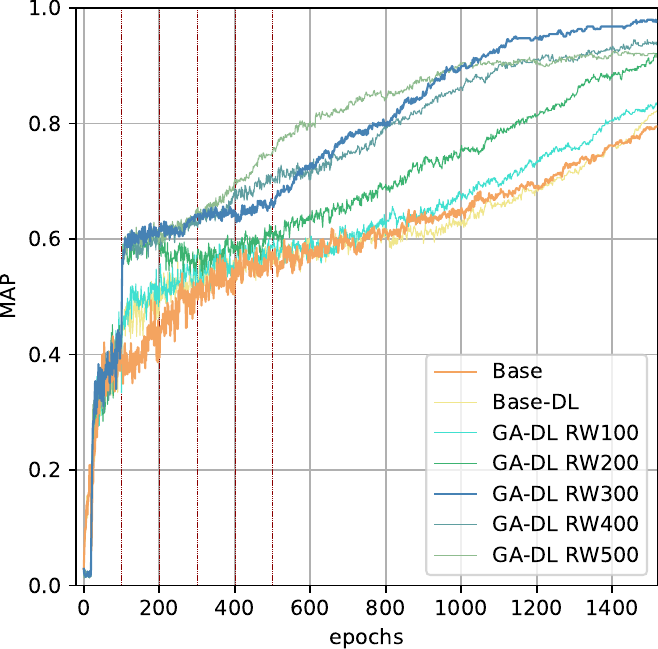}}
\caption{Performances of our algorithm under different hyperparameter settings on the synthetic tree. Base denotes the baseline algorithm, GA is our algorithm where DL is the dilation operation, RW is the re-weighting strategy followed by the threshold epoch number $N_{\mathrm{tc}}$.}
\label{fig: syn result}
\end{figure*}

We synthesize a balanced tree $T$ of 5 layers including the root node. Each non-leaf node in $T$ has 5 children. The edges are directed, pointing from children to parents. We illustrate our algorithm on this synthetic dataset.

We compare our algorithm (Algorithm \ref{alg:G-Aalgorithm}) to the baseline \Poincare embedding algorithm \citep{Nickel2017}. In both algorithms, we set the dimension to be $2$, learning rate to be $0.5$, batch size to be $50$, and the number of negative samples $m$ to be $50$. 

The comparison between learning procedures is presented in Figure \ref{fig: syn process}. As demonstrated, our algorithm learns visually more balanced embeddings with less illness. Quantitative results of the comparison, including the number of illness, MAP and MR, are presented in Figure \ref{fig: syn tc weight} and Figure \ref{fig: syn result}. We compare different $N_{\mathrm{tc}}$ in Figure \ref{fig: syn result} and different $\eta_{\mathrm{tc}}$ in Figure \ref{fig: syn tc weight}. Our algorithm produces less illness than the baseline algorithm and achieves the highest MAP and MR. 

Figure \ref{fig: syn tc weight} shows our explorations on how to use transitive closure edges.They tend to increase the overall effective gradient magnitude and draw vertices of a same subtree tightly together. Therefore, this could help reduce capacity illness quickly (See Figure \ref{fig: syn tc weight} (a)) and eliminate inter-subtree illness (See Figure \ref{fig: syn tc weight} (c)). However, it might also confuse the ground-truth tree edges with the added ones, thus increasing intra-subtree illness (See Figure \ref{fig: syn tc weight} (b)). When we assign weights to the transitive closure edges, this side effect is mitigated: $\eta_{tc} = 0.2$ yields the best results.

Figure \ref{fig: syn result} shows our explorations on how to use dilation and reweighting. These two operations should be applied  after certain epochs of training so that 1) the subtrees are pushed relatively far from each other \footnote{A similar idea is shown in the burn-in stage of \citep{Nickel2017}.} to ensure the dilation operation will push vertices in the appropriate directions, and 2) before the vertices are already pushed to places with sufficient capacity. Empirically we find this threshold epoch number $N_{tc} = 300$ yields the best results.

\newpage

\section{Real-World Dataset Experiments}\label{sec: real world experiments}
\subsection{Dataset Statistics}\label{sec: dataset statistics}
Table \ref{table: statistics} displays statistics of several datasets used in our experiments (including the Yelp Challenge Dataset \footnote{\url{https://www.yelp.com/dataset/documentation/main}}, the Commodity Catalog Dataset, and the WordNet dataset \citep{miller1998wordnet}). We report the number of nodes $|\mathcal{V}|$, the number of edges $|\mathcal{E}|$, maximum degree, variance of all degrees, and tree depths. We make subjective remarks to their size and characteristics. 

\begin{table}[h]
	\centering
	\begin{tabular}{c | ccccccc}
	\toprule\toprule
	Dataset & $|\mathcal{V}|$ & $|\mathcal{E}|$ & Max deg & Std of deg & Depth & Size & Characteristics \\
	\cmidrule{1-8}
	Yelp Challenge (T) & 1587 & 1586 & 194 & 7.9 & 4 & small & shallow \\
	Commodity Catalog (T) & 134812 & 134811 & 726 & 26.4 & 6 & large & shallow; bushy \\
	WordNet Verbs (T) & 13542 & 13541 & 361 & 9.1 & 12 & medium & deep \\
    WordNet Nouns (T) & 82115 & 82114 & 666 & 6.6 & 15 & large & deep \\
    WordNet Nouns (C) & 82115 & 769130 & 82114 & 913.6 & 15 & large & deep; dense \\
    \bottomrule\bottomrule
\end{tabular}
\caption{Dataset statistics (T: Tree; C: Closure)}
\label{table: statistics}
\end{table}

\subsection{Yelp Challenge Dataset (Tree) Results} \label{sec:yelp challenge tree}

The results for Yelp Challenge (Tree) are illustrated in Table \ref{table: yelp}. In all algorithms, we set the learning rate to be 1.0, batch size to be 10, and the number of negative samples $m$ to be 50.
\begin{table}[!h]
	\centering
	\begin{tabular}{c c c c c c c c}
	\toprule\toprule
	& & Dim & MR & MAP & Capacity & Intra & Inter\\
	\cmidrule{1-8}
	\multirow{15}{*}{\rotatebox[origin=c]{90}{\parbox{2.4cm}{\centering YELP CHALLENGE \textbf{Reconstruction}}}} & \multirow{1}{*}{\textbf{Euclidean}} & \multirow{5}{*}{2} & 27.612 & 0.107 & 1440 & 82 & \textbf{28}\\
	& \multirow{1}{*}{\textbf{\citet{Nickel2017}}} &  & 1.681 & 0.855 & 292 & \textbf{45} & 64\\
	& \multirow{1}{*}{\textbf{\citet{nickel2018learning}}} &  & 1.520 & 0.894 & 163 & 47 & 74\\
	& \multirow{1}{*}{\textbf{GA-DL (Ours)}} &  & 1.351 & \textbf{0.926} & \textbf{86} & 65 & 47\\
	& \multirow{1}{*}{\textbf{GA-DL-RW (Ours)}} &  & \textbf{1.202} & 0.914 & 162 & 54 & 37\\
	\cmidrule{2-8}
	\cmidrule{2-8}
	\cmidrule{2-8}
	& \multirow{1}{*}{\textbf{Euclidean}} & \multirow{5}{*}{5} & 11.320 & 0.326 & 1236 & 9 & 44 \\
	& \multirow{1}{*}{\textbf{\citet{Nickel2017}}} &  & 1.063 & 0.988 & \textbf{1} & \textbf{0} & 28\\
	& \multirow{1}{*}{\textbf{\citet{nickel2018learning}}} &  & 1.101 & 0.984 & \textbf{1} & 11 & 26\\
	& \multirow{1}{*}{\textbf{GA-DL (Ours)}} &  & 1.062 & 0.987 & 3 & 9 & 20\\
	& \multirow{1}{*}{\textbf{GA-DL-RW (Ours)}} &  & \textbf{1.030} & \textbf{0.989} & 3 & 9 & \textbf{18}\\
	\cmidrule{2-8}
	\cmidrule{2-8}
	\cmidrule{2-8}
	& \multirow{1}{*}{\textbf{Euclidean}} & \multirow{5}{*}{10} & 1.528 & 0.910 & 183 & 1 & 14 \\
	& \multirow{1}{*}{\textbf{\citet{Nickel2017}}} &  & 1.042 & 0.990 & \textbf{0} & \textbf{0} & 25\\
	& \multirow{1}{*}{\textbf{\citet{nickel2018learning}}} &  & 1.064 & 0.987 & \textbf{0} & 9 & 23\\
	& \multirow{1}{*}{\textbf{GA-DL (Ours)}} &  & 1.051 & 0.989 & 1 & 5 & 21\\
	& \multirow{1}{*}{\textbf{GA-DL-RW (Ours)}} &  & \textbf{1.015} & \textbf{0.993} & 5 & 6 & \textbf{9}\\
    \bottomrule\bottomrule
\end{tabular}
\caption{Yelp Challenge Dataset}
\label{table: yelp}
\end{table}

\newpage
 
\subsection{WordNet Verbs Dataset (Tree) Results} \label{sec:wordnet verbs tree}
 The results for WordNet Verbs (Tree) are illustrated in Table \ref{table:wordnet verbs}. In all algorithms, we set the learning rate to be 1.0, batch size to be 10, and the number of negative samples $m$ to be 50.
\begin{table}[!h]
	\centering
	\begin{tabular}{c c c c c c c c}
	\toprule\toprule
	& & Dim & MR & MAP & Capacity & Intra & Inter\\
	\cmidrule{1-8}
	\multirow{15}{*}{\rotatebox[origin=c]{90}{\parbox{2.4cm}{\centering WORDNET VERBS \textbf{Reconstruction}}}} & \multirow{1}{*}{\textbf{Euclidean}} & \multirow{5}{*}{2} & 66.302 & 0.124 & 7366 & 3778 & 1879\\
	& \multirow{1}{*}{\textbf{\citet{Nickel2017}}} &  & 8.088 & 0.521 & 4826 & 2582 & 1683\\
	& \multirow{1}{*}{\textbf{\citet{nickel2018learning}}} &  & 6.875 & 0.448 & 5459 & \textbf{1769} & 2812\\
	& \multirow{1}{*}{\textbf{GA-DL (Ours)}} &  & 8.319 & 0.510 & 5023 & 2361 & 1908\\
	& \multirow{1}{*}{\textbf{GA-DL-RW (Ours)}} &  & \textbf{3.559} & \textbf{0.563} & \textbf{4543} & 2741 & \textbf{1399}\\
	\cmidrule{2-8}
	\cmidrule{2-8}
	\cmidrule{2-8}
	& \multirow{1}{*}{\textbf{Euclidean}} & \multirow{5}{*}{5} & 21.245 & 0.295 & 10383 & \textbf{1273} & \textbf{277}\\
	& \multirow{1}{*}{\textbf{\citet{Nickel2017}}} &  & 2.329 & 0.854 & 235 & 2321 & 454\\
	& \multirow{1}{*}{\textbf{\citet{nickel2018learning}}} &  & 2.389 & 0.853 & 285 & 2277 & 468\\
	& \multirow{1}{*}{\textbf{GA-DL (Ours)}} &  & 2.195 & \textbf{0.878} & \textbf{211} & 1443 & 920\\
	& \multirow{1}{*}{\textbf{GA-DL-RW (Ours)}} &  & \textbf{1.722} & 0.841 & 378 & 1306 & 1893\\
	\cmidrule{2-8}
	\cmidrule{2-8}
	\cmidrule{2-8}
	& \multirow{1}{*}{\textbf{Euclidean}} & \multirow{5}{*}{10} & 6.240 & 0.677 & 4205 & 1424 & 300\\
	& \multirow{1}{*}{\textbf{\citet{Nickel2017}}} &  & 1.950 & 0.856 & 232 & 2322 & 457\\
	& \multirow{1}{*}{\textbf{\citet{nickel2018learning}}} &  & 1.945 & 0.855 & 257 & 2338 & \textbf{441}\\
	& \multirow{1}{*}{\textbf{GA-DL (Ours)}} &  & \textbf{1.654} & \textbf{0.884} & \textbf{184} & 1456 & 828 \\
	& \multirow{1}{*}{\textbf{GA-DL-RW (Ours)}} &  & \textbf{1.654} & 0.842 & 241 & \textbf{1255} & 2110\\
    \bottomrule\bottomrule
\end{tabular}

\caption{WordNet Verbs}
\label{table:wordnet verbs}
\end{table}

\subsection{WordNet Verbs Nouns (Tree) Results} \label{sec:wordnet nouns tree}

The results for WordNet Nouns (Tree) are illustrated in Table \ref{table:wordnet nouns}. In all algorithms, we set the learning rate to be 1.0, batch size to be 50, and the number of negative samples $m$ to be 50.

\begin{table}[!h]
 	\centering
 	\begin{tabular}{c c c c c c c c}
 	\toprule\toprule
 	& & Dim & MR & MAP & Capacity & Intra & Inter\\
 	\cmidrule{1-8}
 	\multirow{10}{*}{\rotatebox[origin=c]{90}{\parbox{2.4cm}{\centering WORDNET NOUNS \textbf{Reconstruction}}}} & \multirow{1}{*}{\textbf{Euclidean}} & \multirow{5}{*}{5} & 151.321 & 0.235 & 46563 & 21942 & 2993\\
 	& \multirow{1}{*}{\textbf{\citet{Nickel2017}}} &  & 71.271 & 0.322 & 48844 & \textbf{17293} & 438\\
 	& \multirow{1}{*}{\textbf{\citet{nickel2018learning}}} &  & 74.625 & 0.230 & 50950 & 18777 & 311\\
 	& \multirow{1}{*}{\textbf{GA-DL (Ours)}} &  & 19.313 & 0.481 & 35119 & 21158 & 444\\
 	& \multirow{1}{*}{\textbf{GA-DL-RW (Ours)}} &  & \textbf{2.869} & \textbf{0.697} & \textbf{13394} & 23907 & \textbf{27}\\
	\cmidrule{2-8}
 	\cmidrule{2-8}
 	\cmidrule{2-8}
 	& \multirow{1}{*}{\textbf{Euclidean}} & \multirow{5}{*}{10} & 23.113 & 0.278 & 23978 & 31018 & 15745\\
 	& \multirow{1}{*}{\textbf{\citet{Nickel2017}}} &  & 41.014 & 0.324 & 49152 & \textbf{17276} & 154\\
 	& \multirow{1}{*}{\textbf{\citet{nickel2018learning}}} &  & 38.395 & 0.241 & 50986 & 18690 & 100\\
 	& \multirow{1}{*}{\textbf{GA-DL (Ours)}} &  & 7.900 & 0.754 & 10647 & 13547 & 5551 \\
 	& \multirow{1}{*}{\textbf{GA-DL-RW (Ours)}} &  & \textbf{2.738} & \textbf{0.722} & \textbf{9483} & 24844 & \textbf{32}\\
     \bottomrule\bottomrule
 \end{tabular}
 \caption{WordNet Nouns}
 \label{table:wordnet nouns}
 \end{table}
 
 \newpage
 
\subsection{Commodity Catalog (Tree) Results} \label{sec:commodity catalog tree}

The results for Commodity Catalog (Tree) are illustrated in Table \ref{table:commodity catalog}. In all algorithms, we set the learning rate to be 1.0, batch size to be 10, and the number of negative samples $m$ to be 50.

\begin{table}[!h]
 	\centering
 	\begin{tabular}{c c c c c c c c}
 	\toprule\toprule
 	& & Dim & MR & MAP & Capacity & Intra & Inter\\
 	\cmidrule{1-8}
 	\multirow{10}{*}{\rotatebox[origin=c]{90}{\parbox{2.4cm}{\centering COMMODITY CATALOG \textbf{Reconstruction}}}}
 	& \multirow{1}{*}{\textbf{Euclidean}} & \multirow{5}{*}{5} & 163.044 & 0.024 & 132276 & \textbf{1480} & 700\\
 	& \multirow{1}{*}{\textbf{\citet{Nickel2017}}} &  & 86.759 & 0.063 & 130479 & 2001 & 174\\
 	& \multirow{1}{*}{\textbf{\citet{nickel2018learning}}} &  & 50.393 & 0.082 & 127437 & 1968 & 238\\
 	& \multirow{1}{*}{\textbf{GA-DL (Ours)}} &  & 5.405 & \textbf{0.745} & \textbf{49837} & 1971 & 301\\
 	& \multirow{1}{*}{\textbf{GA-DL-RW (Ours)}} &  & \textbf{2.951} & 0.683 & 56922 & 7711 & \textbf{26}\\
 	\cmidrule{2-8}
 	\cmidrule{2-8}
 	\cmidrule{2-8}
 	& \multirow{1}{*}{\textbf{Euclidean}} & \multirow{5}{*}{10} & 59.133 & 0.052 & 131806 & \textbf{1676} & 289\\
 	& \multirow{1}{*}{\textbf{\citet{Nickel2017}}} &  & 67.283 & 0.071 & 130212 & 1998 & 147\\
 	& \multirow{1}{*}{\textbf{\citet{nickel2018learning}}} &  & 36.836 & 0.112 & 124366 & 1964 & 196\\
 	& \multirow{1}{*}{\textbf{GA-DL (Ours)}} &  & 2.167 & \textbf{0.978} & \textbf{1871} & 1861 & 390 \\
 	& \multirow{1}{*}{\textbf{GA-DL-RW (Ours)}} &  & \textbf{2.015} & 0.881 & 19902 & 5525 & \textbf{26}\\
     \bottomrule\bottomrule
 \end{tabular}

 \caption{Commodity Catalog}
 \label{table:commodity catalog}
 \end{table}
 
 The Commodity Catalog (Tree) dataset is generated from real-world e-commerce data. It is extremely bushy. According to \citet{sala2018representation} the hyperbolic space is capable to embed even extremely bushy trees. However, we find the baseline algorithm in \citet{Nickel2017} can not fully exert such capability. This is because a bushy tree requires large local capacity, but the baseline algorithm takes very long time to reach such capacity and instead easily gets stuck at local optimum. Experimentally, the baseline algorithm does not perform well on such bushy dataset, while our algorithm with simple dilation yields significantly better results \footnote{Noticeably, since the tree is bushy, adding transitive closure edges would largely increase intra-class illness, which makes MAP drop.}.
 
\subsection{WordNet Nouns (Closure) Results} \label{sec:wordnet nouns closure}

The results for WordNet Nouns (Closure) are illustrated in Table \ref{table:wordnet nouns closure}. In all algorithms, we set the learning rate to be 1.0, batch size to be 50, and the number of negative samples $m$ to be 50. \footnote{Results of \citet{Nickel2017} are based on their official implementation and hyperparameters in \url{https://github.com/facebookresearch/poincare-embeddings}.}

\begin{table}[!h]
	\centering
	\begin{tabular}{c c c c c c c c}
	\toprule\toprule
	& & Dim & MR & MAP & Capacity & Intra & Inter\\
	\cmidrule{1-8}
	\multirow{4}{*}{\rotatebox[origin=c]{90}{\parbox{2.4cm}{\centering WORDNET NOUNS CLOSURE \textbf{Reconstruction}}}}
	&&&&&&& \\
	& \multirow{1}{*}{\textbf{\citet{Nickel2017}}} & 
	10 & 4.736 & 0.772 & 46192 & 151324 & \textbf{9942}\\ \\
	& \multirow{1}{*}{\textbf{GA-DL (Ours)}} & 10 & 4.788 & 0.781 & 41756 & 144358 & 11987\\ \\
	& \multirow{1}{*}{\textbf{GA-DL-RW (Ours)}} & 10 & \textbf{4.270} & \textbf{0.797} & \textbf{38277} & \textbf{134541} & 12575\\
    \bottomrule\bottomrule
\end{tabular}
\caption{WordNet Nouns Closure}
\label{table:wordnet nouns closure}
\end{table}